\documentclass[letterpaper, 10 pt, conference]{ieeeconf}  

\IEEEoverridecommandlockouts                              

\usepackage{dblfloatfix}

\usepackage{graphics} 
\usepackage{epsfig} 
\usepackage{mathptmx} 
\usepackage{times} 
\usepackage{amsmath} 
\usepackage{amssymb}  
\usepackage[normalem]{ulem}

\usepackage{hyperref}
\usepackage[capitalize]{cleveref}

\usepackage[
    style=ieee,
    sorting=none,
    backend=biber,
    maxbibnames=4,
    doi=false,     
    url=false,     
    eprint=false,  
    isbn=false     
]{biblatex}
\title{\LARGE \bf
Compliant Sphere Lattice Contact: Distributed Contact Modeling for Sphere-Based Robot Representations
}

\author{Nataliya Nechyporenko, Ava Abderezaei, Alessandro Roncone}

\begin{document}

\maketitle
\thispagestyle{empty}
\pagestyle{empty}


\begin{abstract}
Contact planning in robotics requires models that are both computationally efficient and physically accurate. Sphere-based robot representations satisfy the first requirement by enabling fast collision checking and differentiable geometry, but sacrifice physical accuracy by relying on point contact which cannot capture contact patch area, pressure distributions, rotational stiffness, or frictional moments. We introduce Compliant Sphere Lattice Contact (CSLC), a distributed contact model that operates natively on sphere representations by modeling the robot interface as a compliant lattice of surface spheres connected through anchor and lateral springs. When pressed against an object, the lattice deforms to produce a spatially distributed contact patch that improves the physical accuracy of sphere-based contact. We validate CSLC across two independent solvers and show preliminary results demonstrating contact patch formation and improved grasp stability.
\end{abstract}

\begin{figure*}[t]
\centering
\includegraphics[width=\textwidth]{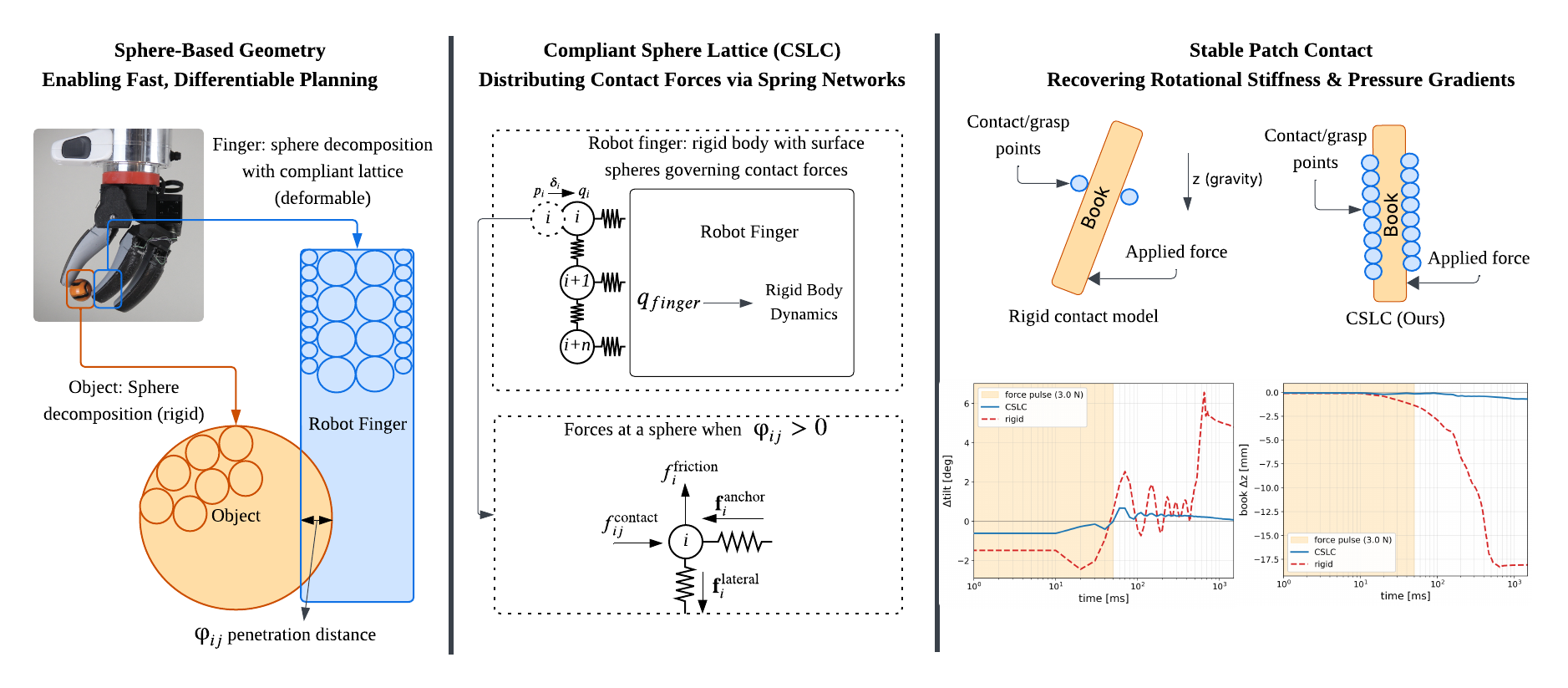}
\caption{Compliant Sphere Lattice Contact (CSLC) overview. (Left) The robot finger and object are represented using sphere primitives for fast, differentiable planning, with the finger modeled as a deformable lattice. (Center) CSLC distributes contact forces by connecting surface spheres through a network of anchor and lateral springs. (Right) Preliminary results of a grasped object subjected to a rotational disturbance. The standard point contact model exhibits instability, whereas CSLC generates a distributed contact patch that recovers rotational stiffness and restoring torque, successfully stabilizing the object.}
\label{fig:diagram}
\end{figure*}

\section{Introduction}

Physical contact is the fundamental mechanism through which robots interact with their environment. To accomplish tasks like washing dishes or clearing a table, robots must accurately predict how objects will respond to contact. Model-based contact formulations enable these predictions for simulation, trajectory optimization, planning, and control \cite{le2024contact}. However, applying these models in real time introduces a challenging computational tradeoff. Contact models must capture complex physical interactions while remaining lightweight enough to evaluate inside tight control loops.

Modern robotics increasingly utilizes sphere primitives as a primary geometric representation to satisfy the computational demands of planning in real time. Motion planning frameworks such as cuRobo \cite{sundaralingam2023curobo, sundaralingam2026curobov2}, VAMP \cite{thomason2024motions}, and AkinoPDF \cite{duong2026ultrafast} represent robot links as collections of spheres because they enable exceptionally fast collision checking and admit analytical signed distance queries. Furthermore, contact planners adopt spheres because their geometry is fully differentiable \cite{kurtz2026inverse, pang2023global}. However, the contact models applied to these sphere representations sacrifice physical accuracy. The standard approach relies on point contact. Point contact inherently lacks a contact patch area, meaning it cannot model rotational stiffness, frictional moments, or pressure distributions \cite{masterjohn2022velocity, elandt2019pressure}. When a robot grasps a flat book or a curved cylinder, a point contact model localizes the entire reaction force at a single infinitely small location. This creates fragile grasps that are highly sensitive to microscopic pose errors. To compensate for this instability, planners are forced to artificially inflate friction coefficients or command excessive grip forces. These workarounds inevitably suppress the nuanced mechanics required for dexterous manipulation and risk damaging fragile objects.

We introduce Compliant Sphere Lattice Contact (CSLC) to fill this gap. CSLC is a distributed contact model that unifies the speed of sphere primitives with the physical fidelity previously available only to methods based on meshes. Rather than relying on rigid points or volumetric intersections, CSLC models the robot interface as a compliant lattice. The surface consists of sphere primitives connected by anchor springs and lateral springs. When pressed against an object, these spheres displace to form a spatially distributed contact patch. With this formulation, we take a step toward capturing physical properties of contact like rotational stiffness and pressure distributions that were previously only available to mesh-based methods. We implement CSLC in the Newton simulator and report preliminary results on gripper-object interactions during grasping.

\section{Related Work}
Most robotics simulators rely on rigid or locally compliant point‑contact models (Linear Complementarity Problem (LCP), Nonlinear Complementarity Problem (NCP), or Cone Complementarity Problem (CCP) variants with Hunt–Crossley‑style normal laws), which regularize force–penetration but still concentrate load at isolated points and therefore cannot represent contact patch area, pressure distributions, or rotational stiffness \cite{le2024contact, drake, todorov2012mujoco, coumans2019, liang2018gpu}. Distributed patch‑based models such as Pressure Field Contact (PFC) and its hydroelastic and discrete variants address these limitations by precomputing pressure fields over volumetric or surface meshes and integrating across a contact surface \cite{elandt2019pressure,masterjohn2022velocity,Cas21}. 

Surface-contact rigid-body methods model extended rigid patches and their equivalent wrenches \cite{Xie16,Xie18}, whereas soft-finger models encode compliance in a local lumped contact law \cite{Xyd99,Cio06,Xu19,Bak14}. CSLC differs from both by representing the interface as a compliant distributed sphere lattice, so patch formation and load sharing emerge directly on sphere-based geometric surfaces.

Complementarity‑free and contact‑implicit formulations instead focus on how to integrate contact into optimization and control, reformulating time‑stepping to remove explicit complementarity constraints and enabling real‑time dexterous manipulation, but they still assume either point‑based or mesh‑based contact laws and do not tackle distributed contact in sphere‑based representations \cite{jin2024complementarity, sleiman2019contact, doshi2019contact, kurtz2022contact}.

In contrast to these existing approaches, our method achieves the physical fidelity of distributed pressure fields directly on sphere-based representations, with a straightforward path to a differentiable formulation. 

\section{Methods}



\subsection{Sphere Lattice Kinematics}

We start with an arbitrary rigid body that has a defined position and
orientation. A collection of surface spheres covers this body. Each
sphere $i$ has a rest position $\mathbf{p}_i$ fixed to the underlying
rigid body, a contact radius $r_i$ representing the thickness of the
compliant skin, and a precomputed outward unit normal
$\hat{\mathbf{n}}_i$ determined by its position on the body surface.
When the body interacts with another object, the skin at sphere~$i$
deforms by a vector displacement
$\boldsymbol{\delta}_i \in \mathbb{R}^3$, shifting the sphere centre
from $\mathbf{p}_i$ to a \emph{deformed centre}
\begin{equation}
    \mathbf{q}_i = \mathbf{p}_i + \boldsymbol{\delta}_i.
    \label{eq:def_centre}
\end{equation}
The contact radius $r_i$ stays fixed and only the sphere centre
moves. The underlying rigid body obeys ordinary rigid-body dynamics
independent of $\boldsymbol{\delta}$. The displacements
$\boldsymbol{\delta}_i$ are the state variables of the contact
problem and encode how the compliant skin deforms under load. We decompose $\boldsymbol{\delta}_i$ in sphere
$i$'s local rest frame as
\begin{equation}
    \delta_{n,i} = \boldsymbol{\delta}_i \cdot \hat{\mathbf{n}}_i,
    \qquad
    \boldsymbol{\delta}_{t,i}
        = \boldsymbol{\delta}_i - \delta_{n,i}\,\hat{\mathbf{n}}_i,
    \label{eq:delta_decomp}
\end{equation}
where $\delta_{n,i} < 0$ is the signed normal compression of the skin and
$\boldsymbol{\delta}_{t,i}$ is tangential shear of the compliant layer
relative to the rigid body. The tangential component carries the
stick-slip restraint of~\cref{sec:friction}; without it the model
would have no representation of static friction prior to macroscopic
slip. We stack the per-sphere displacements into a state vector
$\boldsymbol{\delta} = [\boldsymbol{\delta}_1^\top, \ldots,
\boldsymbol{\delta}_n^\top]^\top \in \mathbb{R}^{3n}$.

\subsection{Forces on Each Sphere}

Four distinct forces act on every surface sphere $i$. These forces determine how the body deforms during contact.

\subsubsection{Anchor Spring}
A spring connects each sphere centre to its rest position and resists
displacement of the compliant skin. Resolving the anchor force in
sphere $i$'s rest frame gives separate stiffnesses on the normal and
tangential axes,
\begin{equation}
    \mathbf{f}_i^{\text{anchor}}
        = -k_a\,\delta_{n,i}\,\hat{\mathbf{n}}_i
          - \rho\,k_a\,\boldsymbol{\delta}_{t,i},
    \label{eq:anchor}
\end{equation}
where $k_a > 0$ is the normal anchor stiffness and
$\rho \in (0,1]$ is the tangent ratio. The isotropic case
$\rho = 1$ recovers
$\mathbf{f}_i^{\text{anchor}} = -k_a\,\boldsymbol{\delta}_i$, while
$\rho = 1/3$ matches the shear modulus of a nearly-incompressible
elastomer.

\subsubsection{Lateral Coupling}
Lateral springs connect adjacent spheres and spread the contact load
across a larger area instead of keeping it concentrated at a single
point. The lateral force on sphere $i$ is
\begin{equation}
    \mathbf{f}_i^{\text{lateral}}
        = -k_\ell \sum_{j \in \mathcal{N}(i)}
          \bigl(\boldsymbol{\delta}_i - \boldsymbol{\delta}_j\bigr),
    \label{eq:lateral}
\end{equation}
where $\mathcal{N}(i)$ is the set of neighbouring spheres and
$k_\ell \geq 0$ controls the load spreading. The coupling acts
component-wise on the vector displacement, so it transmits both normal
compression and tangential shear between neighbours.

The lateral spring coupling plays the role of a continuous pressure
field as in the hydroelastic model~\cite{masterjohn2022velocity}. Both
mechanisms ensure that the contact load is distributed across the
patch rather than concentrated at a single point.

\subsubsection{Contact Force}

In our formulation, we assume that both the gripper lattice and the
target object are represented as a collection of spheres. However, by
sampling points on a mesh and using those points in place of spheres,
we can also apply CSLC to mesh based objects. This allows us to
benchmark CSLC against mature mesh based solvers such as MuJoCo. We
write $\mathbf{t}_j$ for the $j$-th element of the target surface, which
is an object sphere when the target is sphere based and a sampled point
when the target is mesh based. Each $\mathbf{t}_j$ carries an outward
normal $\hat{\mathbf{n}}_{\text{face},j}$ and an area element $A_j$, and
these areas sum to the total target surface area. When the target is
sphere based, the normal is the sphere's outward radial direction, the
area element is its packing area, and the face penetration described
below then reduces to an overlap between two spheres. When the target
is mesh based, the normal is the surface normal at the sampled point
and the area element is the Voronoi area of that point.

For each lattice sphere $i$ that reaches $\mathbf{t}_j$, the contact
direction is the face normal $\hat{\mathbf{n}}_{ij} =
\hat{\mathbf{n}}_{\text{face},j}$, and the signed face penetration
overlap is
$\phi_{ij} = -\hat{\mathbf{n}}_{\text{face},j} \cdot
(\mathbf{q}_i - \mathbf{t}_j)$,
which is zero at first geometric contact between the lattice sphere
centre and the target's tangent plane and grows monotonically with
penetration depth. The per-pair contact force on lattice sphere $i$
from target element $j$ follows a Hertz-like compliance law,
\begin{align}
\mathbf{f}_{ij}^{\text{contact}} &= k_c\,A_j\,s_{ij}\,
        \phi_{ij}^{\text{eff}}\,\hat{\mathbf{n}}_{ij},
    \label{eq:contact} \\
\phi_{ij}^{\text{eff}} &= \max(\phi_{ij},\,0)^{3/2},
    \label{eq:phi_eff} \\
s_{ij} &= \frac{w_{t,ij}}{\sum_k w_{t,kj}},
    \label{eq:share}
\end{align}
with contact stiffness $k_c$ (units $\mathrm{N/m^{2/7}}$) and locality
kernel $w_{t,ij}$ that marks each lattice sphere reaching target
element $\mathbf{t}_j$ within its tangent-plane radius $r_i$. The share
weight $s_{ij}$ gives each reaching lattice sphere a fraction of the
target's area element $A_j$, so each target element contributes $A_j$
of area to the total force regardless of how many lattice spheres reach
it. The total body force
$\mathbf{F}^{\text{body}} = \sum_{i,j} \mathbf{f}_{ij}^{\text{contact}}$
is therefore a Riemann approximation of the surface integral
$\int_\Omega k_c\,\phi^{\text{eff}}\,\hat{\mathbf{n}}_{\text{face}}\,dA$
analogous to the pressure integral in pressure-field
methods~\cite{elandt2019pressure} but with a Hertz-like growth. Force
is zero at $\phi_{ij} \leq 0$ and grows smoothly with penetration, with
a $C^1$ onset at $\phi_{ij}=0$. In the future, we plan to smooth the
$\max(\cdot,0)$ clamp for gradient-based use.

\subsubsection{Pre-sliding Friction}
\label{sec:friction}
The compliant skin develops tangential shear before macroscopic slip
occurs. Following the pre-sliding (elastoplastic) friction
literature~\cite{de1995new, Cas23}, we tie friction to the tangential
displacement $\boldsymbol{\delta}_{t,i}$ from \cref{eq:delta_decomp}
rather than to a sliding velocity. Unlike LuGre-style models, which
carry a separate state variable for the bristle deflection,
$\boldsymbol{\delta}_{t,i}$ is already part of the lattice's
displacement field and emerges from the quasistatic equilibrium
solve. Let $f_{n,i} = \lvert \mathbf{F}_i^{\text{contact}} \cdot
\hat{\mathbf{n}}_i \rvert$ denote the aggregate normal-axis magnitude
of the contact force on sphere $i$. The friction force is
\begin{equation}
    \mathbf{f}_i^{\text{friction}}
        = -\frac{k_{\text{stick}}\,\mu\,f_{n,i}}
                {k_{\text{stick}}\,\lVert\boldsymbol{\delta}_{t,i}\rVert
                 + \mu\,f_{n,i}}\,
          \boldsymbol{\delta}_{t,i},
    \label{eq:friction}
\end{equation}
where $k_{\text{stick}}$ is the tangential stick stiffness and $\mu$
the Coulomb coefficient. The expression interpolates smoothly between
an elastic spring $-k_{\text{stick}}\,\boldsymbol{\delta}_{t,i}$ at
small shear and Coulomb saturation at magnitude $\mu\,f_{n,i}$. Being
quasistatic and tangentially stateless, it captures pre-slip
elasticity and the Coulomb limit but not kinetic sliding dynamics,
which suffices for the grasping regime we target.

\subsection{Contact Stiffness}
In a uniform flat contact each
engaged sphere transmits load through three series elements. These are
the anchor $k_a$, the contact element, and the target modulus
$k_e^{\text{target}}$. Linearising the contact law at a reference
penetration $\phi_0$, one sphere contributes a tangent stiffness
$\kappa_c = \tfrac{3}{2}\,k_c\,A_c\,\sqrt{\phi_0}$, where $A_c$ is its
contact area from \cref{eq:contact}. With $N_{\text{contact}}$ such
chains in parallel, matching the aggregate to a target bulk stiffness
$k_e^{\text{bulk}}$ gives
\begin{equation}
    \frac{1}{\kappa_c}
      = \frac{N_{\text{contact}}}{k_e^{\text{bulk}}}
        - \frac{1}{k_a}
        - \frac{1}{k_e^{\text{target}}},
    \qquad
    k_c = \frac{2\,\kappa_c}{3\,A_c\,\sqrt{\phi_0}}.
    \label{eq:calibration}
\end{equation}
This relation grounds $k_c$ in a measurable bulk modulus and can be
inverted offline for tuning, however, to simplify the analysis in this work we set $k_c$ directly.

\subsection{Quasistatic Equilibrium Solver} \label{sec:solver}

At each timestep the lattice displacement field $\boldsymbol{\delta}$
is found by balancing the four forces on every surface sphere $i$,
\begin{equation}
    \mathbf{f}_i^{\text{anchor}}
    + \mathbf{f}_i^{\text{lateral}}
    + \mathbf{F}_i^{\text{contact}}
    + \mathbf{f}_i^{\text{friction}}
    = \mathbf{0},
    \label{eq:qstatic}
\end{equation}
with the four terms supplied by
\cref{eq:anchor,eq:lateral,eq:contact,eq:friction}.
We solve it with a damped Jacobi sweep, warm-started from the
$\boldsymbol{\delta}$ of the previous timestep. Updates are
independent across spheres and map naturally to GPU parallelism.

After the lattice solve converges, CSLC hands the deformed sphere
centres $\mathbf{q}_i$ and the per-sphere contact and friction forces
to the host rigid-body simulator, which advances the body state
according to its own integration scheme. CSLC is agnostic to the
choice of simulator, and we have integrated it with both a MuJoCo rigid-body pipeline (mesh-sampled) and a position-based dynamics pipeline (sphere/particle).

\section{Experiments, Results \& Discussion}

\subsection{Lattice Deflection Under Point Load}
\begin{figure}[t]
    \centering
    \includegraphics[width=\columnwidth, trim=0 0pt 0 0, clip]{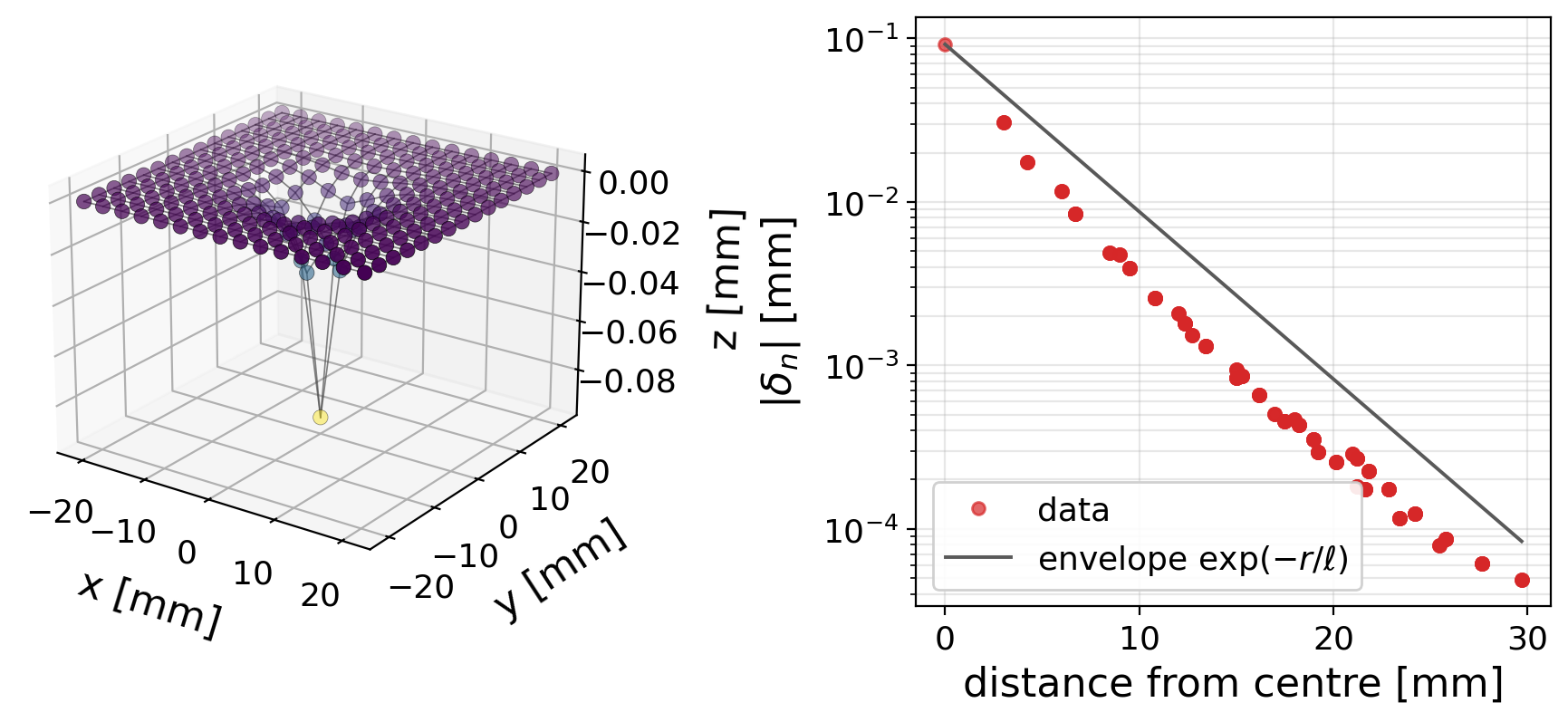}
\caption{Deflection of a flat $15\times15$ CSLC lattice under a single contact. The left panel shows the deformed lattice, where only
the center sphere receives a contact force of $58$\,mN and the
surrounding deformation is produced entirely by graph-Laplacian lateral
coupling. The right panel shows that the radial compression
$|\delta_n|$ decays on the scale $\ell = h\sqrt{k_\ell/k_a} = 4.24$\,mm,
consistent with the lattice's screened-Poisson Green's function. The
grey curve is the exponential envelope $\exp(-r/\ell)$ that bounds its
large $r$ asymptote.}
\label{fig:deflection}
\end{figure}
We first verify that the spring network distributes a localised load as
intended before coupling it to rigid-body dynamics. We build a
$15\times15$ flat lattice with $3$\,mm spacing, $k_a = 100$\,N/m,
$k_\ell = 200$\,N/m, and $k_c = 10^{9}$\,N/m$^{7/2}$, and press a single
contact sample into the centre sphere from above. Solving the
quasistatic equilibrium of \cref{eq:qstatic} gives the deformation in
\cref{fig:deflection}. Only the centre sphere receives a direct contact
force, because every other sphere lies outside the sample's tangential
locality kernel, so the surrounding deformation is produced entirely by the
graph-Laplacian lateral coupling.

Away from the load, the compression decays with a characteristic
length $\ell = h\sqrt{k_\ell/k_a}$ set by the lattice spacing $h$ and
the lateral-to-anchor stiffness ratio. For our parameters this gives
$\ell = 4.24$\,mm. The right panel of \cref{fig:deflection} confirms
the decay, with the data tracking the predicted $\exp(-r/\ell)$
envelope. The patch width is therefore controlled entirely by the
$k_\ell/k_a$ ratio.

\subsection{Stable Lifting Near the Coulomb Limit}

\begin{figure}[t]
\centering
\includegraphics[width=0.15\textwidth]{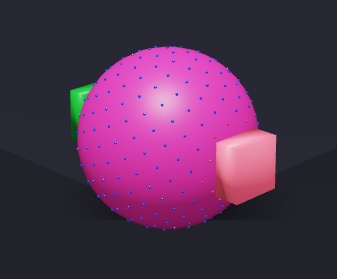}
\hfill
\includegraphics[width=0.15\textwidth]{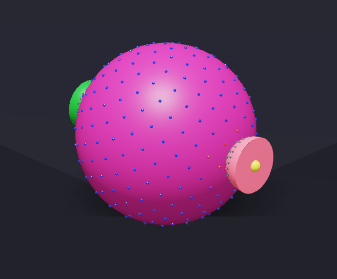}
\hfill
\includegraphics[width=0.15\textwidth]{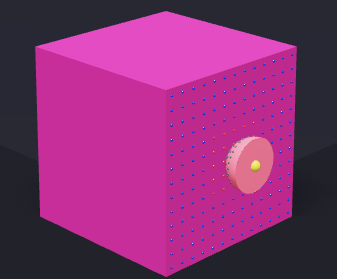}
\caption{The three grasp configurations from \cref{fig:lift} in MuJoCo,
shown left to right as a flat pad on a sphere, a dome pad on a sphere,
and a dome pad on a box. The markers on each object are the CSLC contact
samples. Their density over the contact region carries the
distributed contact patch that enforces the grip.}
\label{fig:grasp_scenes}
\end{figure}

\begin{figure}[t]
\centering
\includegraphics[width=0.235\textwidth]{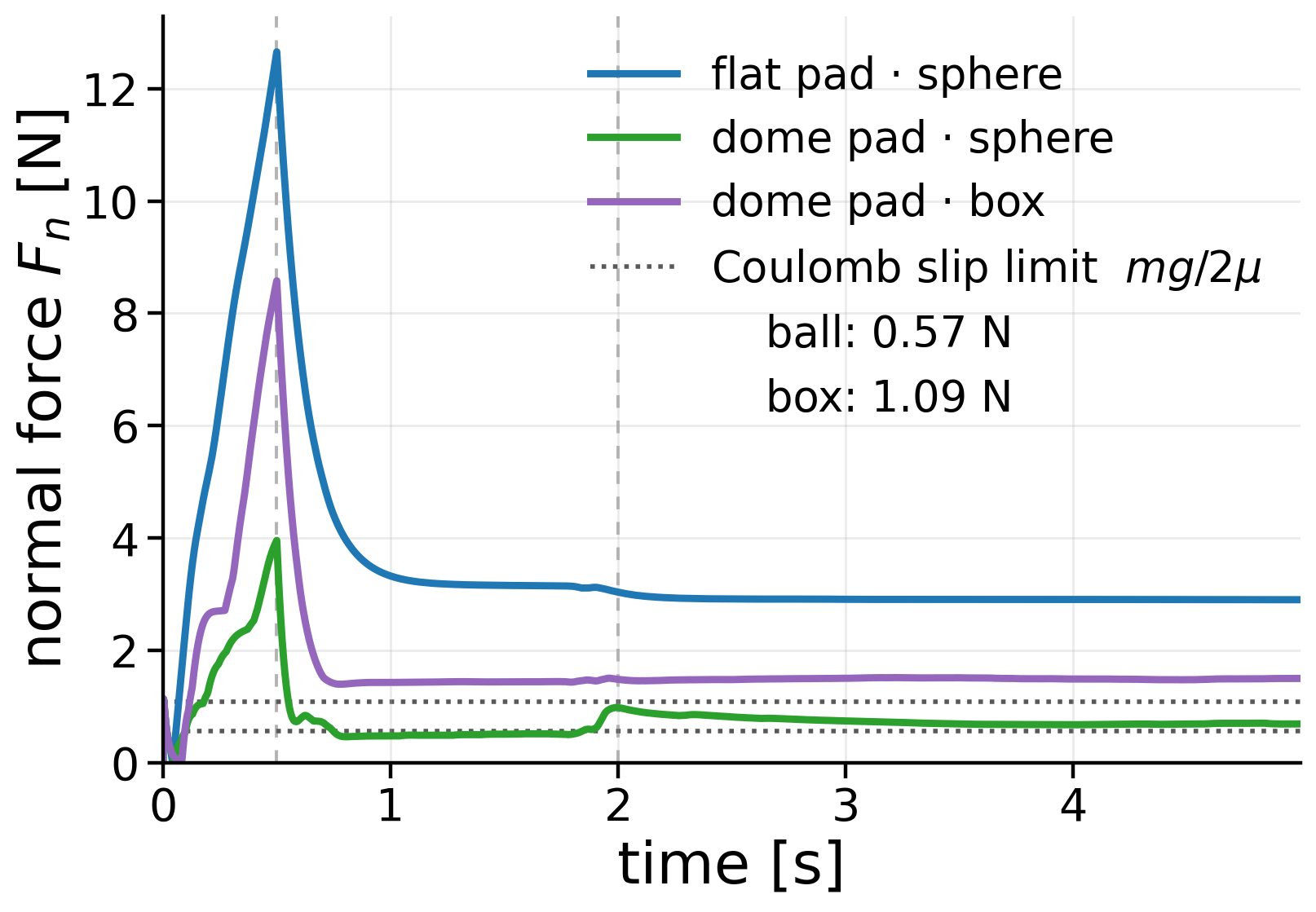}
\hfill
\includegraphics[width=0.235\textwidth]{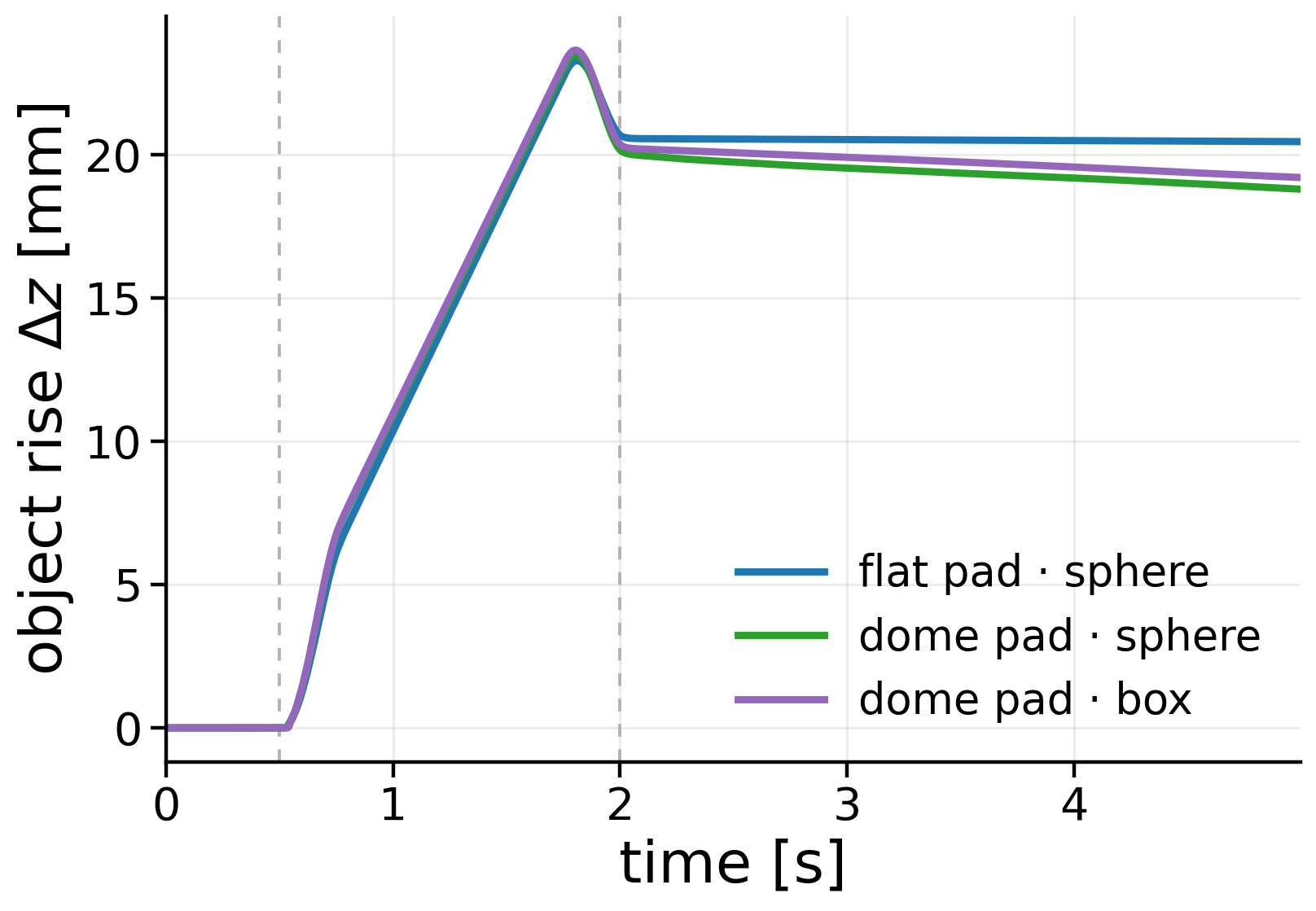}
\caption{CSLC integrated into a MuJoCo squeeze, lift, and hold grasp
for three pad and object pairs. The left panel shows the per-pad normal
force with the dotted Coulomb slip floors $mg/2\mu$, and the right panel
shows the object rising about $20$\,mm and maintaining the hold.}
\label{fig:lift}
\vspace{-1em}
\end{figure}

To test CSLC for grasping applications, we integrate it into a MuJoCo
pipeline and run a two-finger squeeze, lift, and hold on the three pad
and object pairs in \cref{fig:grasp_scenes}, namely a flat pad on a
sphere, a dome pad on a sphere, and a dome pad on a box. The flat pad
presents a broad planar lattice and the dome pad a small convex one, so
for the same squeeze (1mm $\phi$) the flat pad engages a larger contact patch while
the dome pad engages a small one.

A grasp holds the object against gravity only if the friction at the
two pads can carry its weight. With friction coefficient $\mu$ and
per-pad normal force $F_n$, this requires
\begin{equation}
    2\mu F_n \ge mg
    \quad\Rightarrow\quad
    F_n \ge \frac{mg}{2\mu},
    \label{eq:coulomb}
\end{equation}
which sets a slip floor $mg/2\mu$ on the grip force. With $\mu = 0.5$
this floor is $0.57$\,N for the ball and $1.09$\,N for the box.

\Cref{fig:lift} shows the results. After the squeeze transient the
normal force settles to a steady plateau for every pair, and the object
rises smoothly to about $20$\,mm and holds without slipping. Because it
engages the smaller patch, each dome grasp settles near its slip floor,
at $0.7$\,N against the $0.57$\,N ball floor and $1.5$\,N against the
$1.09$\,N box floor, roughly $1.4$ times the limit, while the broader
flat patch holds the ball more firmly at $2.9$\,N. CSLC therefore
produces stable grasps in the low-force regime just above the Coulomb
slip limit, which is the gentle grip that a distributed compliant patch
is meant to enable.

\subsection{Rotational Grasp Stability}
We integrate CSLC into the PBD-R solver from our prior
work~\cite{abderezaei2026physically}, then grasp a rectangular box that simulates a rigid book
($2\times10\times20$\,cm, $0.3$\,kg) with two curved
fingertips approximated with sphre via \cite{nechyporenko2025morphit}. The fingers then lift the book into free space, and then we apply a force pulse ($3$\,N,
$50$\,ms) that tries to rotate the book out of the grasp. We compare
CSLC against a rigid point-contact baseline, the same fingertip treated
as a rigid convex shape, with the scene, friction ($\mu=1.0$), and grip
kinematics held fixed so that only the contact model varies. The
squeeze normal force is identical for both models, giving a friction
capacity roughly two orders of magnitude above the book's weight, so
the grasp is never translationally friction-limited and the test
isolates rotational stiffness. As \cref{fig:diagram} shows, the rigid
point contact cannot generate a restoring couple, so the book pitches
out of the grasp with over $6$ degrees of tilt and slides $18$\,mm down,
whereas CSLC bounds the response to $\pm0.7$ degrees with negligible
height loss and damps back to equilibrium. The distributed patch
supplies this restoring torque through its off-axis compliant springs,
recovering the rotational stiffness that point contact lacks while
operating entirely on sphere primitives.

\section{Conclusion} \label{sec:conclusion}
CSLC assumes quasistatic lattice relaxation, so the solver breaks down
during high-speed impacts where the skin cannot equilibrate within a
timestep. It is also more expensive than point contact at runtime,
costing roughly three times as much per step on the same grasp scene.
Part of this gap is intrinsic, since CSLC resolves a distributed
contact patch where point contact resolves only a few points, and each
step solves a quasistatic equilibrium over the lattice rather than a
single force evaluation per contact pair.
Two changes would address these limits directly. A better-conditioned
solve with fewer Jacobi iterations would narrow the runtime gap, and a
dynamic extension beyond the quasistatic assumption would let the model
handle impacts. Separately, the contact law currently relies on a hard
$\max(\cdot,0)$ clamp and hard active-set culls in the forward path, and
replacing them with smooth surrogates would make CSLC fully
differentiable, opening it to gradient-based planning and control.
This work is our initial step toward the broader goal of creating a
contact model that is differentiable, physically informative, and
practical for real-time robotics. These properties are essential for
model-based manipulation and learning-based methods, whose performance
is directly tied to the strengths and limitations of the underlying
contact models~\cite{yang2025physics, kanehira2025rl}.




\AtNextBibliography{\footnotesize}
\printbibliography

\end{document}